# The Nature of Mathematical Modeling and Probabilistic Optimization Engineering in Generative AI

Fulu Li
Contact: fulu@alum.mit.edu

## Abstract

In this paper, we give an in-depth analysis on the mathematical problem formulations and the probabilistic optimization explorations for some of the key components in Transformer model [33] in the field of generative AI. We explore and discuss some potential further enhancement for current state of the art methods for some key underlying technologies of generative AI models from algorithmic and probabilistic optimization perspective. In particular, we present an optimal solution for sub-word encoding (SWE) based on similar initial settings as that of byte-pair encoding (BPE) algorithm in [9] with similar objectives as that of WordPiece approach in [28, 31] to maximize the likelihood of the training data. We also present cross entropy optimization method to optimize hyperparameters for word2vec model [17]. In addition, we propose a factored combination of rotary positional encoding (RoPE) [32] and attention with linear biases (ALiBi) [23] with a harmonic series. We also present a probabilistic FlashAttention [6, 7] (PrFlashAttention) method with a probability distribution over block distances in the matrix to decide which block is likely to participate in a given round of attention computation while maintaining the lower triangle shape of the tensor for autoregressive language models by re-shaping the tensors. Finally, we present staircase adaptive quantization (SAQ) of key-value (KV) cache for multi-query attention (MQA) based on the framework presented in [16] to have gradual quantization degradation while achieving reasonable model quality and cost savings.

## 1. Introduction

In the year of 1953, Albert Einstein, arguably the most notable scientific giant in the $20^{th}$ century, once commented on the development of Western Science and he concluded that "the development of Western Science in based on two great achievements: the invention of the formal logical system by Greek philosophers, and the discovery of the possibility to find out *causal relationships* by systematic experiments (Renaissance)". Large language models (**LLMs**) made great strides or breakthroughs with deep neural networks via unsupervised learning in recent years mainly due to the discovery of the possibility to find out *causal attention* (essentially a paradigm shift for the study of applied science without the need to find out clear causal relationship) by Transformer models [33] with massive data sets and massive parallel computations with graphics processing units (GPUs) or tensor processing units (TPUs).

Mathematically, a language model is a framework and/or mechanism to compute the joint probability and/or conditional probability of natural language texts for some given language [21, 34]. In this paper, we start with the Transformer model [33] and give an in-depth analysis on the mathematical problem formulations and the probabilistic optimization approaches for some of the key components in the Transformer model in generative AI. We explore and discuss some potential further enhancement for current state of the art methods for some key underlying technologies of generative AI models from algorithmic and probabilistic optimization engineering perspective.

Notably, maximum-likelihood estimation is often used as the de facto mathematical tool for parameter estimation such as the weights and bias in feed-forward network (**FFN**) or multi-layer perceptron (**MLP**) in the blocks/layers of Transformer model [33]. For *non-independent* variables, which is often the case in practice, maximizing log likelihood to minimize the cost or loss function with or without constraints in machine learning can be unsolvable in closed form. That is probably why we often have to resort to using *iterative* procedures to train large language models (**LLMs**) in epochs, updating the weights and bias values of the neurons/nodes in deep neural networks after each round of belief propagation.

Gradient descent, which was first proposed by Augustin-Louis Cauchy in 1847 [13], is a mathematical method for unconstrained optimization. More precisely, it is a first-order *iterative* approach to minimize a differentiable multivariate objective function such as the cost or loss function in machine learning. The basic idea is to take repeated steps in the opposite direction of the gradient of the function at the current point, which is the direction of the steepest descent. Gradient descent method is very useful in machine learning for minimizing the cost or loss function as the objective function to optimize. The convergence properties of gradient descent method for non-linear optimization problems were first studies by Haskell Curry in 1944 [5]. As an extension of gradient descent, stochastic gradient descent, which uses mini-batch sampling during the optimization process, serves as the most basic algorithm used for training most deep neural networks today [12, 21].

In this paper, we analyze and explore the mathematical problem formulations and probabilistic optimization engineering aspects for some of the most critical components of large language models (LLMs) underpinning today's generative AI, in particular the Transformer model [33]. The major contributions of the paper include:

a) We present a systematic analysis on the mathematical problem formulations and probabilistic optimization explorations for some of the major components of Transformer model [33] underpinning today's generative AI;
b) We explore further enhancement with algorithmic and probabilistic optimization solutions for the state of the art approaches in the related areas: in particular, we present an optimal solution for sub-word encoding (SWE) based on similar initial settings as that of byte-pair encoding (BPE) algorithm [9] with similar objectives as that of WordPiece approach [28, 31] to maximize the likelihood of the training data.
c) We also present cross entropy optimization method to optimize hyperparameters for word2vec model in [17].
d) In addition, we propose a factored combination of rotary positional encoding (RoPE) [32] and attention with linear biases (ALiBi) [23] with a harmonic series.
e) We also present a probabilistic FlashAttention [6, 7] (PrFlashAttention) method with a probability distribution over block distances in the matrix to decide which block is likely to participate in a given round of attention computation while maintaining the lower triangle shape of the tensor for autoregressive language models by re-shaping the tensors.
f) Finally, we present staircase adaptive quantization (SAQ) of key-value (KV) cache for multi-query attention (MQA) based on the framework presented in [16] to have gradual quantization degradation while achieving reasonable model quality and cost savings.

The rest of the paper is organized as follows: We discuss related notions behind some engineering practices in generative AI in Section 2. We gave an in-depth analysis on mathematical modeling and probabilistic optimization engineering of LLM in generative AI in Section 3. We discuss some further enhancements for pre-training and post-training of LLM in Section 4. The summary and future directions are given in Section 5.

## 2. Related Notions Behind Some Engineering Practices

In probability theory, the law of large numbers states that the average of the results obtained from a large number of independent random samples converges to the true value. As mentioned earlier, in the training of deep neural networks during the maximum likelihood estimation of those parameters, i.e., the weights and bias of those neural nodes in the deep neural network, stochastic gradient descent method is used in the iterative optimization process. According to the law of large numbers, the gradients computed from these mini-batches of training data set are expected to fluctuate around the true gradient for the whole training data set. Therefore, the **mini-batch gradient** on average is expected to indicate an adequate direction for changing the parameters during stochastic gradient descent optimization process [5, 12, 13, 21].

In probability theory and statistics, variance is the expected value of the squared deviation from the mean of a random variable. **Layer normalization** in the Transformer model [33] is used for regularization, where the mean and variance of the parameters in each layer are computed in order to make sure that the resulting outputs of the given layer have a well-behaved distribution, with the value of expectation/mean at 0.0 and the value of variance at 1.0 [21].

Depending on the real application scenarios, most of the generative pre-trained transformer (**GPT**) models adopted **decoder-only Transformer** model [19], where the query matrix of $Q$ and the key matrix of $K$ are all of full rank due to the autoregressive nature of contiguous lower triangle shape along the diagonal in the corresponding matrix during attention computation. A matrix is of full rank if its rank is the same as its smaller dimension. Notably, a matrix of full rank can carry more information that a matrix of the same size with lower rank, where some rows or columns in the matrix are not of linear independence. That is probably why decoder-only transformer models become the state of the art in natural language generation, in particular for generative AI.

The logistic sigmoid function also has its root in probability theory and it was used as an activation function in deep neural networks, where the value of the logistic function is in the range of (0, 1) and can be interpreted as a probability. Another popular activation function for deep neural networks, named rectified linear unit (**ReLU**), achieves better gradient propagation when training deep neural networks. Non-linearity is essential for neural networks as it allows the algorithm to deduce complex patterns in the data. Non-linearity is accomplished by activation functions [30]. In retrospect, there was the argument that single layer perceptron can not solve XOR classification if there is a linear activation function for the neural network, which speaks for itself about the importance of the introduction of non-linear activation functions to deep neural networks today. ReLU was first used by Alston Householder in 1941 as a mathematical abstraction of biological neural networks [11].

In probability theory, the output of the softargmax function can be used to represent a categorical distribution, where a probability distribution is over k different possible outcomes. For the **softmax** function is the Transformer model [33], it is essentially a normalized exponential function that converts the vector representation in the neural networks back to the probability distribution over the tokens, which facilitates next token prediction for generative AI.

Now let us take a look at the rationale behind some **emergent ability** of some generative AI model when the size of the language model, i.e., the number of parameters of the deep neural networks underpinning the given language model, is large enough and the size of the training data for the given neural networks is large enough. Let $\mu$ be the number of times that event $A$ occurs during $n$ independent trials and let $p$ denote the probability that event $A$ occurs for every trial. According to Bernoulli's law of large numbers [2], for any positive real number of $\epsilon$, we have

$$\lim_{n->\infty} \Pr\{ |\frac{\mu}{n} - p| < \epsilon \} = 1 \tag{1}$$

An interesting observation is that when the number of parameters of the language model in the deep neural networks is large enough and the size of the training data for the neural networks is large enough, the event that some emergent ability for a given generative AI model may appear is of high possibility, which could be the new form of law of large numbers in the field of large language models (LLMs) for generative AI.

## 3. Mathematical Modeling of LLM for Generative AI

In the following, we discuss the probabilistic optimization formulations for autoregressive language model that is widely used in GPT-like models (generative pre-trained transformer) [19], pre-training of the foundation model [21], fine-tuning of the model with additional domain knowledge of (question, answer) pairs [25], reinforcement learning with human feedback (**RLHF**) [20, 27], direct preference optimization (**DPO**) [24] and identity preference optimization (**IPO**) [1] for alignment in generative AI.

Mathematically, **autoregressive language model** (**ALM**) generating a sequence of text by predicting next token given previous tokens is based on conditional probability, where the objective of the model is to estimate the joint probability of the given sequence of tokens. According to the chain rule of probability, we have:

$$\Pr(s) = \Pr(s_1, s_2, \ldots, s_T) = \prod_{t=1}^{T} \Pr(s_t | s_1, s_2, \ldots, s_{t-1}) \tag{2}$$

where the token sequence of $s$ is given by $s = (s_1, s_2, \ldots, s_T)$, $T$ is the length of the token sequence, the model predicts the probability of each token of $s_t$ based on all the previous tokens of $s_1, s_2, \ldots s_{t-1}$ in the sequence [21, 34].

The **pre-training process of LLM** with deep neural networks such as the Transformer model [33] is to minimize the cross-entropy loss of the training data set, i.e., the sum of the negative log-likelihood of of true tokens at each position in the sequence of the training data set, and we have the probabilistic optimization objective as follows:

$$\min_{\theta}( - \sum_{t=1}^{T} \log \Pr(s_t | s_1, s_2, \ldots, s_{t-1}; \theta)) \tag{3}$$

where $\theta$ is the model parameters, $Pr()$ is the probability distribution over the vocabulary [3, 34], the function of softmax() in the Transformer mode converts a vector of numbers into a vector of probabilities, where the probability of $\Pr(s_t | s_1, s_2, ..., s_{t-1})$ is the probability of the true label token index at position $t$ given all preceding tokens of $s_1, s_2, ..., s_{t-1}$.

On the other hand, after the pre-training of the foundation model, the model is further fine-tuned with additional domain knowledge of (question, answer) pairs, the probabilistic optimization objective of **fine-tuning** is to maximize the log-likelihood of the correct answers of the additional training data set as follows:

$$\max_{\theta}(\sum_{i=1}^{N} \log \Pr(A_i | Q_i; \theta)) \tag{4}$$

where $\theta$ is the model parameters, $Q_i$ and $A_i$ is the $i^{th}$ question and answer pair respectively [25], $N$ is the number of total number of (question, answer) pairs, $\Pr(A_i | Q_i)$ is the probability of the correct answer of $A_i$ given question of $Q_i$.

After the fine-tuning of the model with domain knowledge of (question, answer) pairs, **reinforcement learning with human feedback** (**RLHF**) is often used to further improve the model's instruction-following capabilities [20]. The basic idea is to construct a reward model of $R(x)$ to estimate the quality of the model's outputs based on pair-wise of (prompt, completion) training data. The policy/language model is optimized based on methods such as proximal policy optimization (**PPO**) [27] and the probabilistic optimization objective is to minimize the cross entropy loss, which incentivizes it to make predictions that are closer to actual human ratings/feedbacks:

$$\min_{\Pi_\theta}(-\mathbb{E}[R(Q, A) - \beta \times \mathrm{KL}(\Pi_\theta(A | Q) \parallel \Pi_{\theta_{old}}(A | Q))]) \tag{5}$$

where $\mathbb{E}[x]$ indicates the expectation value of $x$ in probability theory, i.e., the first moment, $Q$ indicates the prompt and $A$ denotes the completion, $\beta$ is a hyperparameter that controls the strength of the Kullback–Leibler (KL) divergence penalty and $\beta > 0$, $\Pi_\theta$ is the current policy/language model, $\Pi_{\theta_{old}}$ is the old policy/language model.

KL divergence is also called relative entropy and mathematically it is defined as:

$$\mathrm{KL}(W \parallel Z) = \sum_{x \in \chi} (W(x) \times \log(\frac{W(x)}{Z(x)})) \tag{6}$$

The combination of RLHF and PPO has led to great success in practice such as InstructGPT and GPT4 [1]. However, as pointed out in [8] that RLHF is often slow and quite unstable in practice, in particular in a distributed learning environment. Direct preference optimization (**DPO**) [24] and its variants such as identity preference optimization (**IPO**) [1] are becoming popular alternative solutions without the rewarding stage, while performing reinforcement learning (RL) to learn the policy with a single maximum likelihood objective [24].

The loss that DPO is trying to optimize, given an empirical data set of $D$, as a function of $\Pi_\theta$, i.e., the policy/language model to optimize, is given by [1]:

$$\min_{\Pi_\theta}(-\mathbb{E}_{(x, y_w, y_l) \sim D}[-\log(\sigma(\tau \log(\frac{\Pi_\theta(y_w | x)}{\Pi_\theta(y_l | x)}) - \tau \log(\frac{\Pi_{ref}(y_w | x)}{\Pi_{ref}(y_l | x)})))]) \tag{7}$$

where $\Pi_{ref}$ denotes some reference policy, $\tau$ is a hyperparameter that controls the strength of the Kullback–Leibler (KL) divergence penalty and $\tau > 0$, $\sigma(.)$ denotes the sigmoid function and plays the role of normalization and the data set of $D$ is given by:

$$D = (x_i, y_{w,i} \succ y_{l,i})_{i=1}^{N} \tag{8}$$

where $x_i$ is a given prompt, $y_{w,i}$ and $y_{l,i}$ are two completions given the prompt of $x_i$, $y_{w,i} \succ y_{l,i}$ indicates the preference of $y_{w,i}$ over $y_{l,i}$.

The key findings in [24, 1] is that when the Bradley-Terry model that represents the preference function as a sigmoid of the difference of rewards perfectly fits the preference data and the optimal reward policy is obtained from preference optimization loss function, then the optimization of **RLHF** objective in Equation (5) perfectly coincides with the optimization of **DPO** objective in Equation (7).

Azar et al in [1] further simplified DPO optimization with a sampled loss for IPO as follows:

$$\min_{\Pi_\theta}(-\mathbb{E}_{(x,y_w,y_l)\sim D}[\tau(\log(\frac{\Pi_\theta(y_w|x)}{\Pi_\theta(y_l|x)}) - \log(\frac{\Pi_{ref}(y_w|x)}{\Pi_{ref}(y_l|x)}) - \frac{\tau^{-1}}{2})^2]) \quad (9)$$

where $\tau$ is a hyperparameter that controls the strength of the Kullback–Leibler (KL) divergence penalty and $\tau > 0$, $\Pi_{ref}$ denotes some reference policy, $\Pi_\theta$ is the policy/language model to optimize, $x$ is a given prompt, $y_w$ and $y_l$ are two completions given the prompt of $x$.

In the following, we focus on some of the widely-used models for generative AI such as the Transformer model [33].

## 3.1. The Transformer Model

As discussed in [33], a Transformer model typically has the following major components: **(a)** a tokenizer that converts a sequence of text into a sequence of tokens; **(b)** an embedding layer that converts tokens and positions of tokens into vector/tensor representations; **(c)** transformer layers that conduct repeated transformations on vector representations via deep learning process, extracting more and more language semantics information, which typically consists of multi-head attention (**MHA**) layer and feedforward network (**FFN**) layer; **(d)** LayerNormalization block that computes the mean and variance of the parameters at each layer such that the outputs of the given layer has a well-behaved distribution with a mean of zero and a variance of 1.0; **(e)** An un-embedding layer that converts the vector representation back to the probability distribution over the tokens with a softmax function; **(f)** Residual connections that bypass one or more layers of neural network computations to improve the flow of gradients during back propagation in order to facilitate deeper neural networks. The identity shortcuts of residual connections essentially skip blocks of layers to preserve features of the propagated signal; **(g)** Next token prediction based on some effective algorithms such as top$-p$ algorithm or top$-k$ algorithm, etc.

In the following, we present some enhancements for some of the state of the art methods for some of the key components in the implementation of the Transformer model [33].

### 3.1.1. Subword Encoding to Maximize the Likelihood of of the Training Data

For the tokenizer, **byte pair encoding** (**BPE**) algorithm [9] is widely used for sub-word encoding to deal with rare and unseen word issues. The basic idea for BPE algorithm is to initialize each word unit/token with one character in the text and greedily merge the *adjacent* pair with the *highest frequency* until the number of remaining tokens/words reaches a given size. However, by only focusing on merged-token frequencies when merging two adjacent pair with the highest frequency may lead to reduced total number of appearances for each word vocabulary item in the training data. Essentially, it is a tradeoff between average word length and the total number of words appearance in the training data while avoiding out-of-vocabulary issues and capturing linguistic meanings as much as possible.

Another widely-used sub-word encoding approach named **WordPiece** [28] algorithm starts by initializing all Unicode characters of the collection for a given language as tokens. Then it combines two tokens (*not* necessarily adjacent) in such a way that the likelihood of the training data is maximized [28]. Song et al presented a fast WordPiece algorithm, whose computational complexity is linear with respect to the input length [31]. Notably, both of BPE [9] and WordPiece [28, 31] methods efficiently addressed the open-ended vocabulary problem by allowing a word to be represented as a sequence of characters if necessary, splitting a word into multiple sub-words. Essentially, subword encoding approaches effectively interpolate between word level inputs for frequent words and character level inputs for rare words [21].

The key differences between BPE [9] and WordPiece [28] are two-folds: (a) the initial settings are different in that BPE initializes each word unit/token with one character in the given input text to learn the word vocabulary while WordPiece initializes each word unit/token with one character among all basic Unicode characters for a given language; (b) the criteria used to merge two word units/tokens are different in that BPE merges two *adjacent* word units/tokens with the highest frequency in the given learning text while WordPiece combines two of the word units/tokens (*not* necessarily adjacent) that maximizes the likelihood of the training data.

Please note that both BPE and WordPiece are *heuristic* greedy algorithms, which make it difficult to judge why the vocabulary size of large language model of $A$, say about 200k, is much larger than that of large language model of $B$, say around 50k, if the training data sets for the two language models are largely the same and both large language models are using decoder-only Transformer model [33] for training of their foundation models. To answer this question, we present an optimal solution for sub-word encoding (SWE) based on $k$-step shortest path algorithm with similar initial-settings to that of BPE as well as an enhanced BPE algorithm (eBPE) to maximize the likelihood of training data.

Following the assumptions in WordPiece [28, 31], we do not focus on word semantics, instead we are more interested in word appearances in the training data set. For the sake of simplicity, we assume that the given learning data set for word vocabulary is the same as the training data set of the language model for BPE. The basic idea is that we follow the logistics of BPE [9] including the initial settings to initialize each word unit/ token with one character in the given learning text and the merging operations of adjacent word units/tokens while **not** based on word frequencies in the given learning text but based on the objective to maximize the likelihood of training data, assuming the learning data set for word vocabulary and the training data set is the same for now. We formulate the optimal sub-word encoding (SWE) problem to maximize the likelihood of the training data with a final number of tokens of $k$ in the learning text as a $k$-step shortest path problem to find the optimal solution, where the number of unique words in the final $k$ tokens is the actual size of the vocabulary.

Usually, for the application of sub-word encoding, the first step is to split the text into words based on spaces and append a special end-of-word symbol such as '_' to each word. This special symbol is important as it marks word boundaries, which prevents the algorithm from confusing the end of one word with the start of another word, in particular for BPE algorithm [9] and its variants.

Borrowing some interesting concepts from Riemann Hypothesis, we consider end-of-word spaces/special end-of-word symbols as *trivial* stops and we consider stops in-between characters within a word as *non-trivial* stops. Assuming we have an initial sequence of individual word unit/tokens, each of which is either a character that belongs to an original word in the learning text/training data or a special end-of-word symbol, say '_', our goal is to merge them into subwords or words based on the original order of those tokens in the sequence such that the likelihood of the training data is maximized. Let $k$ be the final number of tokens/ words, where the number of unique words in the final $k$ tokens/words is the actual size of the vocabulary, let $w$ stand for the number of original words in the learning text/training data, let $k_i$ be the number of sub-words/ tokens/steps within the $i^{th}$ original word in the learning text/training data, let $w_i$ stand for the $i^{th}$ original word in the learning text/training data, let $w_{i,j}$ denote the $j^{th}$ sub-word of the the $i^{th}$ original word in the learning text/training data, let the symbol of " ∘" indicate the concatenation of the sub-words, and we have

$$\sum_{i=1}^{w} k_i \le k \qquad (10)$$

$$k_i \ge 1, \quad i \in [1,...,w] \qquad (11)$$

$$k \ge w \qquad (12)$$

$$w_i = w_{i,1} \circ \ldots \circ w_{i,k_i} \qquad (13)$$

Condition (10), Condition (11) , Condition (12) and Equation (13) indicate that all word units/tokens in the initial sequence of learning text/training data are covered in the $k$-step path from the beginning of the sequence to the end of sequence and trivial stops /end-of-word symbols must be visited once and only once, where each original word has at least one step/one token. We also need to emphasize that the final number of tokens of $k$ for the learning text/training data must be greater than or equal to the number of original words in the learning text/training data.

Please note that to find out the total number of word appearances in the learning text/training data, we need to first find out the unique words in the current tokens. Let $u$ be the number of unique words in the current tokens, let $u_i$ stand for the $i^{th}$ unique word in the current tokens, let $a_{u_i}$ denote the number of appearances for the word of $u_i$ in the learning text/training data, let $n_{total}$ be the total number of word appearances in the learning text/training data, and we have

$$n_{total} = \sum_{i=1}^{u} a_{u_i} \qquad (14)$$

Let $s_i$ stand for the corresponding token/word of the $i^{th}$ step in the $k$-step path from the beginning of the sequence to the end of sequence in the learning text/training data, let $a_{s_i}$ denote the number of appearances for the word of $s_i$ in the learning text/training data, let $c_i$ be the cost of the $i^{th}$ step in the $k$-step path, which can be defined as follows:

$$c_i = \frac{1}{a_{s_i} + 1} \tag{15}$$

Notably, we are using the inverse of the sum of the number of corresponding-word appearances for the corresponding step and one as the cost of a given step to avoid the zero appearance issue. For a $k$-step path, let $c_p$ be the total cost of a given path, which can be defined as follows:

$$c_p = \sum_{i=1}^{k} c_i \tag{16}$$

Based on the cost definition of a step in a given path and the cost definition of a given path in Equation (15) and Equation (16), we can apply Bellman-Ford shortest path algorithm to find the $k$-step shortest path that satisfies Condition (10), Condition (11), Condition (12) and Equation (13) as the optimal solution to maximize the likelihood of the training data.

While there is some good justification for the optimal solution of the $k$-step shortest path that satisfies given Conditions in (10)~(13) based on Bellman-Ford shortest algorithm to maximize the likelihood of of the training data, it may not be an ideal solution for an online tokenization algorithm due to its complexity of $O(kN^3)$, where $k$ is the number of steps and $N$ is the input length in terms of the number of characters in the learning text/training data. In the following, we present an enhanced byte-pair encoding (eBPE) algorithm in which two adjacent token pair are merged where the number of merged-token/word appearances in the learning text/training data is the highest. We repeat this merging operation process until the final number of tokens in the learning text/training data reaches the value of $k$.

More formally, based on similar initial settings as that of byte-pair encoding (BPE) algorithm in [9] with similar objectives as that of WordPiece approach in [28, 31] to maximize the likelihood of the training data, we have the optimal solution for sub-word encoding (SWE) based on Bellman-Ford shortest path algorithm to maximize the likelihood of the training data as follows in Figure 1:

1. Initialize each word unit/token with one character in the input text and mark each end-of-word with a special symbol.
2. Initialize the value of the final number of tokens of $k$.
3. Compute pair-wise step cost based on Equation (15).
4. Apply the relaxation procedures of Bellman-Ford shortest path algorithm to find the $k$-step shortest path from the beginning of the token sequence to the end of the token sequence based on the input text and pair-wise cost functions (Equation (15)) and compute the cost of the path based on Equation (16).
5. Choose the the $k$-step shortest path that satisfies Conditions (10) , Condition (11), Condition (12) and Equation (13).
6. Output the final $k$ tokens
7. The unique tokens/words in the final $k$ tokens constitute the vocabulary from the learning text/training data.

**Figure 1**: An Optimal Solution of Sub-word Encoding (**SWE**) based on Bellman-Ford Shortest Path Algorithm to Maximize the Likelihood of the Training Data.

Based on similar initial settings as that of byte-pair encoding (BPE) algorithm in [9] with similar objectives as that of WordPiece approach in [28, 31] to maximize the likelihood of the training data, we have the enhanced byte-pair encoding (eBPE) algorithm to maximize the likelihood of the training data as follows in Figure 2:

1. Initialize each word unit/token with one character in the input text and mark each end-of-word with a special symbol.
2. Initialize the value of the final number of tokens of $k$.
3. Compute the number of merged-token appearances in the learning text/training data for any two adjacent tokens in the current token sequence.
4. Merge two adjacent token pair whose number of merged-token/ word appearances in the learning text/training data is the highest.
5. Repeat Step 3 and Step 4 until the final number of tokens in the learning text/training data reaches the value of $k$.
6. Output the final $k$ tokens
7. The unique tokens/words in the final $k$ tokens constitute the vocabulary from the learning text/training data.

**Figure 2**: An Enhanced Byte-Pair Encoding (**eBPE**) Algorithm to Maximize the Likelihood of the Training Data.

### 3.1.2. Optimization of Hyperparameters for Word2vec Approach

**Word2vec** [17] is widely used as the state of the art model for obtaining vector representations of words, where the objective for these vectors is to capture information about the meaning of the word based on the surrounding words as much as possible such that those vectors can be useful for predicting the surrounding words in a sentence or a document. As stated in [17], more formally, the objective of the skip-gram model is to maximize the average log probability as follows:

$$\frac{1}{T}\sum_{t=1}^{T}\sum_{-c\leq j\leq c, j\neq 0}\log(\Pr(w_{t+j}\,|\,w_t)) \tag{17}$$

where $T$ is the sequence length of a word sequence of $w_1, w_2, w_3, \ldots, w_T$, $c$ is the size of the training context window. For skip-gram model, it is trying to predict surrounding words with a window of radius of $c$ for every word in the window. Continuous bag of words (CBOW) and skip-gram are two of the most popular frameworks used in word2vec model to get word embeddings [17]. For continuous bag of words (CBOW) model, it is trying to predict the middle word with the input of its surrounding words within a given window. The basic idea is that word vectors are positioned in the vector space in such a way that words that share some common contexts, i.e., close to each other, in the text corpus are located close to one another in the vector space. In the following discussions, we only focus on the skip-gram model (Equation (17)) but the principle can be equally applicable to continuous bag of words (CBOW) model as well.

As pointed out in [14] that much of the superior performance of word2vec [17] in downstream tasks, compared with other similar approaches, is mainly due to the choice of specific hyperparameters such as the size of the training context window, i.e., the hyperparameter of $c$, the dimensionality of the vector, i.e., the hyperparameter of $d$, the word frequency threshold for sub-sampling of high-frequency words, i.e., the hyperparameter of $s_f$, the word count threshold to be considered as part of the training vocabulary for low-frequency word, i.e., the hyperparameter of $m_c$.

Existing hyperparameter optimization methods include: (a) grid search, i.e., essentially an exhaustive search over a given hyperparameter space; (b) random search, which can explore much more potential values than grid search for continuous hyperparameters; (c) Bayesian optimization, which employs a probabilistic model between hyperparameter values and objectives evaluated on a given data set and it has shown to outperform both grid search and random search with fewer evaluations; (d) gradient-based optimization, which computes the gradients with respect to hyperparameters when applicable for some learning algorithms and then optimizes the hyperparameters based on gradient decent philosophy; (e) evolutionary optimization, which employs an evolutionary algorithm to find the best possible hyperparameters for a given learning algorithm; (f) population-based methods, which updates the hyperparameters as well as the weights of neural networks during the training of the models with multiple independent learning processes with different hyperparameteres of the learning algorithm; (g) early stopping-based approach, which starts as a random search over the hyperparameters space and successively prunes low-performing ones until there is only one model remaining.

Notably, it is a combinatorial optimization problem for the selection of the size of the training context window of $c$, the dimensionality of the vector of $d$, the word frequency threshold for sub-sampling of high-frequency words of $s_f$, the word count threshold to be considered as part of the training vocabulary for low-frequency word of $m_c$, for the hyperparameters of word2vec algorithm. Therefore, it is not practical to use some common approach like grid search for the selection of hyperparamters for word2vec model/algorithm.

In the following, we present a probabilistic optimization framework with cross entropy optimization method for the selection of hyperparameters in word2vec [17] model/algorithm.

The basic idea of cross entropy (CE) method [26] is to translate the deterministic optimization problem into a corresponding stochastic one and then use rare event simulation techniques to find the optimal solution. As discussed in [26], cross entropy (CE) method differs from other well-known random search algorithms for global optimization such as simulated annealing, tabu search, and genetic algorithms, which are local search heuristics and employ the notion of local neighborhood structures. Cross entropy (CE) method employs multi-extremal optimization process based on Kullback-Leibler cross-entropy, importance sampling, etc. Therefore, cross entropy (CE) method represents a *global* random search procedure rather than a *local* one.

Our work differs from existing hyperparameter optimization methods in that we use cross-entropy optimization with a probability distribution over hyperparameter space, where some rare event simulation techniques are used to find the optimal values of the hyperparameters for a given learning algorithm with a given training data set. The cross-entropy operation is achieved by virtue of probability update after the performance evaluation after each round of randomly-generated hyperparameter samples based on probability distributions, where a small portion of top performers are given higher probabilities for the next round. The iterative optimization process ends when the convergence conditions are achieved or a pre-defined early-stop criteria is met.

We also need to emphasize that for very large training data set, we can randomly choose a small subset of the training data set for cross entropy optimization for the selection of the hyperparameters of word2vec model/algorithm, then we use the selected hyperparameters to train the model with the whole training data set.

We have some basic notations for the cross entropy optimization of hyperparameters in word2vec model/algorithm in [17] as follows: Let $c$ be the hyperparameter for the size of the training context window. Let $d$ stand for the hyperparameter for the dimensionality of the vector. Let $s_f$ denote the word frequency threshold for high-frequency word for sub-sampling. Let $m_c$ stand for the minimum count threshold for low-frequency word to be considered as part of the training vocabulary. Let $F(c, d, s_f, m_c)$ stand for the performance metric of trained word2vec model for a given tuple value of $(c, d, s_f, m_c)$, a given training data set for word similarity and/or analogy tasks, etc. The details of some related performance metric can be found in [14]. We also assume that the proper value range of $c$ is given as $[a, b]$, where $b > a$, $a, b, c$ are all integers and $a, b, c \in \mathbb{Z}$. We also assume that the proper value range of $d$ is given as $[d_1, d_2]$, where $d_2 > d_1$, $d, d_1, d_2$ are all integers and $d, d_1, d_2 \in \mathbb{Z}$. We also assume that the proper value range of $s_f$ is given as $[f_1, f_2]$, where $f_2 > f_1$, $s_f, f_1, f_2$ are all positive real numbers and $s_f, f_1, f_2 \in \mathfrak{R}^+$. We also assume that the proper value range of $m_c$ is given as $[m_1, m_2]$, where $m_2 > m_1$, $m_c, m_1, m_2$ are all integers and $m_c, m_1, m_2 \in \mathbb{Z}$. Let $\gamma_t$ stand for the benchmark value of $F(c, d, s_f, m_c)$ for the $t^{th}$ round and we have its definition as follows:

$$\gamma_t = \min\{f : \Pr_{(c,d,s_f,m_c)_{t-1}} (F(c, d, s_f, m_c) \geq f) \geq \rho\} \quad (18)$$

where the purpose of the hyperparameters optimization of word2vec model/algorithm is to maximize the performance metric for word similarity and/or analogy tasks, etc. of the word2vec model, $\rho$ normally takes a value of 0.01 so that the event of obtaining high performance is not too rare, the tuple value of $(c, d, s_f, m_c)$ is randomly chosen based on the probability distribution of each variable of $(c, d, s_f, m_c)$ respectively, $(c, d, s_f, m_c)_{t-1}$ represents the set of randomly generated values of $(c, d, s_f, m_c)$ in the $(t-1)^{th}$ round. Essentially, $\gamma_t$ is the top $\rho$-quantile of the performers of the randomly generated $(c, d, s_f, m_c)$ values in the $t^{th}$ round.

As discussed in [26], there are several ways to set the termination conditions. Normally, if for some $t \geq l$, say $l = 5$, and we have

$$\gamma_t = \gamma_{t-1} = \ldots = \gamma_{t-l} \tag{19}$$

then we stop the hyperparameter optimization process for word2vec model/algorithm for the given training data set.

Let $M$ stand for the number of sample tuple values of $(c, d, s_f, m_c)$ in a given round, let $(c, d, s_f, m_c)_i$ denote the $i^{th}$ randomly generated sample tuple value of $(c, d, s_f, m_c)$, $H_{\{\}}$ is an indicator function, $q_i$ is the probability that the given $i^{th}$ tuple value of $(c, d, s_f, m_c)_i$ is being chosen and $q_i$ is initialized as zeros at the beginning. We assume uniform distributions of the values for $(c, d, s_f, m_c)$ in the ranges of $[a, b]$, $[d_1, d_2]$, $[f_1, f_2]$, $[m_1, m_2]$, respectively. For example, randomly choosing a value of $c$ in a given range of $[a, b]$ with uniform distribution is like generating a random number in the range of $[a, b]$ with uniform distribution. The updated value of $q_i$ can be estimated as:

$$q_i^e = \frac{H_{\{F((c,d,s_f,m_c)_i) \geq \gamma\}}}{\sum_{k=1}^{M} H_{\{F((c,d,s_f,m_c)_k) \geq \gamma\}}} \tag{20}$$

While there are solid theoretical justifications for Equation (20), we refer interested readers to [26] for the details.

In order to have a smoothed update procedure, normally we have

$$q_i^t = \alpha \times q_i^e + (1 - \alpha) \times q_i^{t-1} \tag{21}$$

where empirically a value of $\alpha$ between 0.4 and 0.9, i.e., $0.4 \leq \alpha \leq 0.9$, gives the best results [26], $q_i^{t-1}$ is the value of $q_i$ in the previous round, $q_i^e$ is the estimated value of $q_i$ based on the performance in the previous round according to Equation (20), $q_i^t$ stands for the value of $q_i$ in the current round.

Let $B$ be the set of tuple values of $(c, d, s_f, m_c)$ whose performance are in the top $\rho$-quantile of the performers of the randomly generated tuple values of $(c, d, s_f, m_c)$ in a given round and we have

$$B = \{(c, d, s_f, m_c)_i \,|\, H_{\{F((c,d,s_f,m_c)_i) \geq \gamma\}}\} \tag{22}$$

A normalized $q_i^t$ for the top $\rho$-quantile of the performers is given as follows:

$$q_i^n = \frac{q_i^t}{\sum_{(c,d,s_f,m_c)_i \in B} q_i^t} \tag{23}$$

Let $s$ be the factor of favorability, i.e., $s = 10$, towards those top performers in a given round and let $N_s$ be the total number of samples chosen from those top performers and we have

$$N_s = s \times M \times \rho \tag{24}$$

For the next round, we have $N_s$ samples that are randomly chosen from those $|B|$ number of tuple values of $(c, d, s_f, m_c)$, where $|B|$ indicates the number of tuple value elements in the value set of $B$. The rest of the samples, i.e., $(M - N_s)$ samples of $(c, d, s_f, m_c)$, are randomly chosen from the ranges of $[a, b]$, $[d_1, d_2]$, $[f_1, f_2]$, $[m_1, m_2]$, respectively with uniform distributions, which can be accomplished by generating $(M - N_s)$ random tuple numbers of $(c, d, s_f, m_c)$ in the range of $[a, b]$, $[d_1, d_2]$ , $[f_1, f_2]$, $[m_1, m_2]$, respectively, with uniform distributions.

In summary, we have the algorithm of cross entropy optimization for the hyperparameters optimization (CEHPO) in word2vec model/algorithm [17] as follows:

For a given training data set and a given set of hyperparameters to optimize within their corresponding value ranges, we have the detailed description of the cross entropy optimization process for hyperparameters optimization in word2vec model/algorithm [17], .i.e., the CEHPO algorithm, in Figure 3. The cross entropy optimization process will stop when the convergence conditions are satisfied or a pre-defined early-stop criteria is met.

1. Set $t = 1$ and initialize $q_i$ as zeros.
2. Generating $M$ random numbers in the range of $[a, b]$ for $c$, in the range of $[d_1, d_2]$ for $d$, in the range of $[f_1, f_2]$ for $s_f$, in the range of $[m_1, m_2]$ for $m_c$, with uniform distribution for the tuple values of $(c, d, s_f, m_c)$ if it is the first time.
3. Generating $(M - N_s)$ random numbers in the range of $[a, b]$ for $c$, in the range of $[d_1, d_2]$ for $d$, in the range of $[f_1, f_2]$ for $s_f$, in the range of $[m_1, m_2]$ for $m_c$, with uniform distribution for the tuple values of $(c, d, s_f, m_c)$ if it is not the first time.
4. $N_s$ is determined by Equation (24).
5. $N_s$ random samples of $(c, d, s_f, m_c)$ are drawn from those top $\rho$ -quantile performers of $(c, d, s_f, m_c)$ in the last round based on probability distribution of $q_i^n$s
6. Calculate $\gamma_t$ according to Equation (18).
7. Update $q_i$ according to Equation (20) and Equation (21).
8. Update normalized $q_i$ according to Equation (23).
9. If for some $t \geq l$, say $l = 5$, such that $\gamma_t = \gamma_{t-1} = \ldots = \gamma_{t-l}$, then stop, otherwise, reiterate from Step 3.

**Figure 3**: the Cross-Entropy HyperParameter Optimization (**CEHPO**) Algorithm.

### 3.1.3. Rotary Positional Embedding (RoPE) and Attention with Linear Biases (ALiBi)

As of today, rotary positional embedding approach (RoPE) presented in [32] is probably the most widely used positional embedding method for Transformer models in generative AI applications, in particular after LlaMA2 adopted this positional embedding approach. The RoPE approach works well in most scenarios but it may have some challenges for input length extrapolation, where sometimes the sequence length for inference is longer than the maximum training sequence length, in particular when the training data set is not large enough. Attention with Linear Biases (ALiBi) method presented in [23] does not add positional embeddings to word embeddings but it adds biases to query-key attention scores with a penalty that is proportional to their distance, which facilitates the extrapolation performance, in particular when the training data set is not large enough and the inductive biases have some significant impact on extrapolation during inference.

Rotary positional embedding method [32] uses Euler's formula in complex analysis to transform the addition operation of the static position encoding proposed in the original Transformer framework [33] into a multiplication operation, where the related parameters naturally become part of the learning process.

Notably, both rotary positional embedding approach of RoPE and ALiBi method incorporate relative position information, where relative position information is used in RoPE for positional embedding calculation and the relative distance information is used in ALiBi for added linear biases for the query-key attention score.

In the following, we propose **a factored combination** of ALiBi and RoPE for Transformer-based generative AI applications in order to obtain the benefits of extrapolation by ALiBi as well as the benefits of RoPE in other application scenarios.

As mentioned earlier, the method of ALiBi [23] does not change the token embeddings, it can be used along with RoPE [32] with some twists due to the fact that a simple combination of ALiBi and RoPE may lead to the signal from relative position information being too strong as both ALiBi and RoPE uses relative position information in their respective approach. In [23], ALiBi uses a geometric series decaying factor as head-specific slope among different heads for multi-head attention (MHA) and it uses a arithmetic series as a function of the distance. The static, non-learned biases works well for input length extrapolation, where sometimes the sequence length for inference is longer than the maximum training sequence length. As described in [23], attention with linear biases (ALiBi) is defined as follows for the attention score of the $i^{th}$ query of $q_i$ in each head:

$$\text{softmax}(q_i K^T + m \,.\, [-(i-1), -(i-2), \ldots, -2, -1, 0]), \tag{25}$$

where $q_i$ is the $i^{th}$ query and $q_i \in \Re^{1\times d}$ and $d$ is the head dimension, the scaled-dot-product is computed by the attention layer in each head for multi-head attention given the first $i$ keys and $K \in \Re^{i\times d}$, $m$ is a hyperparameter of head-specific slope that is picked before training. The proposed values of $m$ follow a geometric series, where for a model with 8 heads, the geometric series is in the form of $\frac{1}{2}, \frac{1}{2^2}, \ldots, \frac{1}{2^8}$. We refer interested readers to [23] for suggestions on how to use the geometric series with other number of heads.

As discussed in [23], their experiments with trainable parameters of slopes of $m$ did not achieve strong extrapolation results. By using the same slopes of a geometric series of $m$ and adding another harmonic series of factors for the added linear biases as follows may lead to reasonable results. We propose to use a harmonic series to add a different scalar factor at different positions in the form of $\frac{1}{1\times 2}, \frac{1}{2\times 3}, \ldots, \frac{1}{i\times(i+1)}$, where the sum of all those scalar factor values is $\sum_{k=1}^{i}(\frac{1}{k} - \frac{1}{k+1}) = \frac{i}{i+1}$. So, Equation (25) can be rewritten as follows:

$$\text{softmax}(q_i K^T + m \,.\, [\frac{-(i-1)}{1\times 2}, \frac{-(i-2)}{2\times 3}, \ldots, \frac{-2}{(i-2)(i-1)}, \frac{-1}{(i-1)\times i}, \frac{0}{i\times(i+1)}]), \tag{26}$$

The intuition is that we may need to properly adjust the linear biases introduced by ALiBi when we use both ALiBi and RoPE for Transformer-based generative AI applications as both ALiBi and RoPE use relative position information in their computation for eventual attention scores. We will evaluate the effectiveness of Equation (26) for ALiBi along with RoPE as one of our future directions for Transformer-based generative AI applications.

## 4. Pre-Training and Post-Training of LLM

In this section, we focus on some techniques to speed up the pre-training of foundation models and to accelerate the inference process after pre-training, in particular on attention computation.

### 4.1. Probabilistic FlashAttetnion

Attention computation is arguably the most critical component of the Transformer model [33]. Due to the quadratic nature of the attention computation complexity, a lot of efforts have been made to speed up the attention computation such as the efforts by OpenAI team in [4]. One widely-used attention computation approach is called FlashAttention in [6, 7]. The essence of FlashAttention method for attention computation in Transformer model for large language model (**LLM**) is the application of tiling by splitting a large matrix into tiles to have a finer granularity to deal with I/O with differentiated memory (HBM, SRAM, etc.) constraint hierarchy in GPU.

FlashAttention [6, 7] is of exact attention computation in its current form. As discussed in [22], fast causal attention computation for sparse FlashAttention can be more efficient. We considers attention computation in Transformer model for LLM in a probabilistic way. The presented probability density function (PDF) with respect to the block/tile distance in the matrix follows a constrained harmonic deduction philosophy. The presented PrFlashAttention *dynamically* and *probabilistically* skips less-related rows/columns in Query/Key (Q/K) matrix along a tensor dimension, say the number of Head dimension of $H$, in the tensor shape of (Batch, Head, Context Length, Head Dimension) during attention computation while supporting causal masks for auto-regressive models by reshaping the tensors.

In the following, we discuss the probabilistic model for the presented approach of PrFlashAttention and how the masks for each row in the query matrix of $Q$ and each column in the key matrix of $K$ are calculated.

The presented probabilistic model for PrFlashAttention is defined as follows:
For a block distance, say, $n$, within a given range, say, $k$, the probability is set as one, otherwise, it follows a harmonic deduction series.

$$f(n) = \begin{cases} 1, & \text{if } \ 0 \le n \le k; \\ \frac{1}{(n-k)(n-k+1)}, & \text{if } \ n > k; \end{cases} \tag{27}$$

Notably, for the second part, when $n > k$, we have the sum of probability series as: $1 - \frac{1}{max(\lceil N \div B_r \rceil, \lceil N \div B_c \rceil) - k}$, where $N$ is the context length, $B_r$ and $B_c$ are block sizes for matrix $Q$ and matrix $K$ respectively.
Let $n_q$ stand for the number of blocks in each row of matrix $Q$, the normalized probability for each row can be defined as:

$$Pr(r_q) = \frac{\sum_{i=1}^{n_q} Pr(i)}{n_q}, \tag{28}$$

where $Pr(i)$ is the probability of the $i^{th}$ block in the $q^{th}$ row of $r_q$.
Similarly, let $n_k$ denote the number of blocks in each column of matrix $K$, the normalized probability for each column can be defined as:

$$Pr(c_k) = \frac{\sum_{i=1}^{n_k} Pr(i)}{n_k}, \tag{29}$$

where $Pr(i)$ is the probability of the $i^{th}$ block in the $k^{th}$ column of $c_k$.
We use a weighted combination of pre-computed row/column probability and a random number to make the selection process **dynamic**:

$$d_q = Pr(r_q) \times w + r \times (1 - w), \tag{30}$$

where $d_q$ is the decision factor for the $q^{th}$ row, $r$ is a random number between 0 and 1, $w$ is the weight between 0 and 1.

$$d_k = Pr(c_k) \times w + r \times (1 - w), \tag{31}$$

where $d_k$ is the decision factor for the $k^{th}$ column, $r$ is a random number between 0 and 1, $w$ is the weight between 0 and 1.
The adjusted **sparsity** value is also a weighted combination:

$$s_{adj} = p_s \times w + \frac{s}{100} \times (1 - w), \tag{32}$$

where $p_s$ is the $s$ percentile value of the normalized row/column probabilities among rows/columns in the matrix, $s$ is the targeted dropping percentage and $w$ is the weight between 0 and 1.

### 4.2. Adaptive Quantization of KV Cache for Multi-Query Attention

Multi-query attention (MQA) has been proposed mostly to speed up the inference process. The basic idea is to keep the original number of heads for query matrix of $Q$ in multi-head attention (MHA) but have only one head for key matrix of $K$ and value matrix of $V$, which means all the $Q$ heads share the same set of $K$ and $V$ heads, where computed keys and values vectors are cached, without the re-computation of the same key and value vectors at each attention block. In general, Multi-query attention (MQA) with key-value (KV) cache has a neutral effect on model quality and training speed, but can greatly speed up the inference.

In the following, we present Staircase Adaptive Quantization (SAQ) for key-value (KV) cache in multi-query attention (MQA) to further alleviate the problem of KV cache based on the framework in [16] by means of gradual quantization degradation to speed up the inference while achieving reasonable model performance. For a $B$-bit integer, the quantization and de-quantization process can be expressed as follows [16]:

$$Q(X) = \lfloor \frac{X - z_X}{s_X} \rceil \quad (33)$$

$$X' = Q(X) \,.\, s_X + z_X \quad (34)$$

where $Q(X)$ indicates the quantized tensor of $X$, $X'$ is the de-quantized tensor of $Q(X)$, $z_X = \min X$ is the zero-point and $s_X = (\max X - \min X)/(2^B - 1)$ is the scaling factor and the symbol of $\lfloor . \rceil$ is the rounding operation.

Notably, key and value cache of newly generated tokens arrive sequentially in time. Following similar settings in [16], during the pre-fill phase, exact (full precision) key and value tensors are passed to the next layers, even though only the quantized KV cache is retained in memory to reduce the memory footprint and to prevent some re-computations in the decoding phase.

We assume that we have $l_{prompt}$ number of key tokens and $l_{prompt}$ number of value tokens in the pre-fill stage. We also assume that a full precision is expressed as 16-bit quantization such as fp16 (float point 16) that is commonly used in the implementation of tensors, so we can have lower quantization choices such as 8-bit quantization, 4-bit quantization, 2-bit quantization, etc. Let $q_n$ be the number of quantization choices and $B_i$ indicate the number of bits for the $i^{th}$ quantization choices and $B_1$ represents the number of quantization bits of the full precision. Since $l_{prompt}$ can be of arbitrary length, in the pre-fill stage we split the sequence of $l_{prompt}$ tokens into $q_n$ segments, i.e., $S_1, S_2, \ldots, S_{q_n}$, each of which corresponds to a different quantization level, with the segment of $S_1$ corresponds to the full precision. For the sake of simplicity, we assume that all of the segments, i.e., $S_1, S_2, \ldots, S_{q_n}$, are of equal sizes, say segment size of $S$. Please note that the size of the segment of $S_{q_n}$, which corresponds to the one with the lowest quantization level, i.e. 2-bit quantization or 1-bit quantization, could be open-ended as the token sequence grows longer and longer unless it is truncated due to the constraint of cache memory. From the token sequence perspective, the quantization level downgrades by half in terms of quantization bits every $S$ tokes, which looks like a staircase. We present the algorithm for Staircase Adaptive Quantization (SAQ) to have gradual quantization degradation for KV cache in both pre-fill stage and decoding stage based on the framework in [16] for multi-query attention (MQA) to speed up the inference. The details of the presented algorithm for Staircase Adaptive Quantization (SAQ) of KV cache for multi-query attention (MQA) in the phase of pre-fill and decoding are described in the Appendix A.

## 5. Summary and Future Directions

With rapid progress in the field of generative AI, it is often a good idea to reflect on some of the fundamental mathematical modeling and probabilistic optimization tools that powered this AI revolution so that we can make further enhancement systematically down the road.

In this paper, we give an in-depth analysis on the mathematical problem formulations and the probabilistic optimization explorations for some of the key components in Transformer model [33] in the field of generative AI. We explore and discuss some potential further enhancement for current state of the art methods for some key underlying technologies of generative AI models from algorithmic and probabilistic optimization perspective. In particular, we present an optimal solution for sub-word encoding (SWE) based on similar initial settings as that of byte-pair encoding (BPE) algorithm in [9] with similar objectives as that of WordPiece approach in [28, 31] to maximize the likelihood of the training data. We also present cross entropy optimization method to optimize hyperparameters for word2vec model in [17]. In addition, we propose a factored combination of rotary positional encoding (RoPE) [32] and attention with linear biases (ALiBi) [23] with a harmonic series. We also present a probabilistic FlashAttention [6, 7] (PrFlashAttention) method with a probability distribution over block distances in the matrix to decide which block is likely to participate in a given round of attention computation while maintaining the lower triangle shape of the tensor for autoregressive language models by re-shaping the tensors. Finally, we present staircase adaptive quantization (SAQ) of key-value (KV) cache for multi-query attention (MQA) based on the framework presented in [16] to have gradual quantization degradation while achieving reasonable model quality and cost savings.

We will conduct extensive experiments for the proposed approaches as one of our future directions.

## Appendix A:

### Pseudo Code for Staircase Adaptive Quantization (SAQ) Algorithm of KV Cache in Multi-Query Attention (MQA)

**Algorithm**: SAQ Pre-fill and Decoding Algorithm

**Parameters**: group size $G$, segment size $S$, quantization options $q_n$, quantization bits $B_i$

**Procedure** Prefill:

**Input**: $X \in \Re^{l_{prompt} \times d}$

$X_K = XW_K,\ X_V = XW_V$

$X_{V_g} = X_V[:l_{prompt} - S],\ X_{V_r} = X_V[l_{prompt} - S:]$

$s_n = l_{prompt} // S$

**if** $s_n \leq q_n$:

$X_{V_{g,i}} = X_V[l_{prompt} - (i+1) \times S : l_{prompt} - i \times S]$ for $i$ in range[1:$s_n - 2$]

$X_{V_{g,s_n-1}} = X_V[:l_{prompt} - (s_n - 1) \times S]$

**else**:

$X_{V_{g,i}} = X_V[l_{prompt} - (i+1) \times S : l_{prompt} - i \times S]$ for $i$ in range[1:$q_n - 2$]

$X_{V_{g,q_n-1}} = X_V[:l_{prompt} - (q_n - 1) \times S]$

**if** $s_n \leq q_n$:

$Q(X_{V_{g,i}})$ = GrpQuant($X_{V_{g,i}}$, dim = token, qbits = $B_{i+1}$, numGroup = $d//G$) for in range[1:$s_n - 1$]

**else**:

$Q(X_{V_{g,i}})$ = GrpQuant($X_{V_{g,i}}$, dim = token, qbits = $B_{i+1}$, numGroup = $d//G$) for in range[1:$q_n - 1$]

$Q(X_{K_g}), X_{K_r}$ = Kquant($X_K$)

KV cache ← $Q(X_{K_g}), X_{K_r}, Q(X_{V_g}), X_{V_r}$

**Return** $X_K, X_V$

**end**

**Procedure** Decoding:

**Input**: KV cache, $t \in \Re^{1\times d}$

$t_Q = tW_Q, t_K = tW_K, t_V = tW_V$

$Q(X_{K_g}), X_{K_r}, Q(X_{V_g}), X_{V_r} \leftarrow$ KV cache

$X_{K_r} = \mathrm{Concat([X_{K_r}, t_K], dim = token)}$

$X_{V_r} = \mathrm{Concat([X_{V_r}, t_V], dim = token)}$

**if** $\mathrm{len(X_{K_r}) == S}$:

$Q(X_{K_r}), _ = \mathrm{Kquant(X_{K_r})}$

$s_n = len(Q(X_{K_g}))//S$

**if** $s_n \le (q_n - 2)$:

$Q(X_{K_g,i}) = \mathrm{Q(X_{K_g})[-i \times S : (i-1) \times S], dim = token}$, for $i$ in range[1:$s_n$]

$X_{K_g,i} =$ GrpDeQuant($Q(X_{K_g,i})$, dim = channel, qbits = $B_{i+1}$, numGroup = $S//G$) for $i$ in range[1:$s_n$]

$Q(X_{K_g,i}) =$ GrpQuant($X_{K_g,i}$, dim = channel, qbits = $B_{i+2}$, numGroup = $S//G$) for in range[1:$s_n$]

**else**:

$Q(X_{K_g,i}) = \mathrm{Q(X_{K_g})[-i \times S : (i-1) \times S], dim = token}$, for $i$ in range[1:$q_n - 2$]

$Q(X_{K_g,q_n-1}) = Q(X_{K_g})[: -(q_n - 2) \times S]$

$X_{K_g,i} =$ GrpDeQuant($Q(X_{K_g,i})$, dim = channel, qbits = $B_{i+1}$, numGroup = $S//G$) for $i$ in range[1: $q_n - 2$]

$X_{K_g,q_n-1} =$ GrpDeQuant($Q(X_{K_g,i})$, dim = channel, qbits = $B_{i+1}$, numGroup = $(s_n - q_n + 2)S//G$)

$Q(X_{K_g,i}) =$ GrpQuant($X_{K_g,i}$, dim = channel, qbits = $B_{i+2}$, numGroup = $S//G$) for in range[1:$q_n - 2$]

$Q(X_{K_g,q_n-1}) =$ GrpQuant($X_{K_g,q_n-1}$, dim = channel, qbits = $B_{q_n}$, numGroup = $(s_n - q_n + 2)S//G$)

**if** $s_n \le (q_n - 1)$:

$Q(X_{K_g}) = \mathrm{Concat([Q(X_{K_g,s_n}), \ldots, Q(X_{K_g,1}, Q(X_{K_r})], dim = token)}$

**else**:

$Q(X_{K_g}) = \mathrm{Concat([Q(X_{K_g,q_n-1}), \ldots, Q(X_{K_g,1}), Q(X_{K_r})], dim = token)}$

$X_{K_r} \leftarrow$ empty tensor

**end**

**if** $\mathrm{len}(\mathrm{X}_{\mathrm{V_r}}) > \mathrm{S}$:

$Q(X_{V_r'}) = \text{GrpQuant}(X_{V_r}[: -R], \text{dim = token, qbits} = B_2, \text{numGroup} = d//G)$

$Q(X_{V_g}) = \mathrm{Concat}([\mathrm{Q}(\mathrm{X}_{\mathrm{V_g}}), \mathrm{Q}(\mathrm{X}'_{\mathrm{V_r}})], \mathrm{dim} = \mathrm{token})$

**if** $(\mathrm{len}(\mathrm{Q}(\mathrm{X}_{\mathrm{V_g}}))\ \%\ \mathrm{S}) = = 0$:

$s_n = len(Q(X_{V_g}))//S$

**if** $s_n \le (q_n - 1)$:

$Q(X_{V_{g,i}}) = \mathrm{Q}(\mathrm{X}_{\mathrm{V_g}})[-\mathrm{i} \times \mathrm{S} : (\mathrm{i} - 1) \times \mathrm{S}], \mathrm{dim} = \mathrm{token},$ for $i$ in range[2:$s_n$]

$X_{V_{g,i}} = \text{GrpDeQuant}(Q(X_{V_{g,i}}), \text{dim = token, qbits} = B_i, \text{numGroup} = d//G)$ for $i$ in range[2:$s_n$]

$Q(X_{V_{g,i}}) = \text{GrpQuant}(X_{V_{g,i}}, \text{dim = token, qbits} = B_{i+1}, \text{numGroup} = d//G)$ for in range[2:$s_n$]

**else**:

$Q(X_{V_{g,i}}) = \mathrm{Q}(\mathrm{X}_{\mathrm{V_g}})[-\mathrm{i} \times \mathrm{S} : (\mathrm{i} - 1) \times \mathrm{S}], \mathrm{dim} = \mathrm{token},$ for $i$ in range[2:$q_n - 2$]

$Q(X_{V_{g,q_n-1}}) = Q(X_{V_g})[: -(q_n - 2) \times S]$

$X_{V_{g,i}} = \text{GrpDeQuant}(Q(X_{V_{g,i}}), \text{dim = token, qbits} = B_i, \text{numGroup} = d//G)$ for $i$ in range[2: $q_n - 1$]

$Q(X_{V_{g,i}}) = \text{GrpQuant}(X_{V_{g,i}}, \text{dim = token, qbits} = B_{i+1}, \text{numGroup} = d//G)$ for in range[2:$q_n - 1$]

**if** $s_n \le (q_n - 1)$:

$Q(X_{V_g}) = \mathrm{Concat}([\mathrm{Q}(\mathrm{X}_{\mathrm{V_{g,s_n}}}), \ldots, \mathrm{Q}(\mathrm{X}_{\mathrm{V_{g,1}}}, \mathrm{Q}(\mathrm{X}_{\mathrm{V_r}})], \mathrm{dim} = \mathrm{token})$

**else**:

$Q(X_{V_g}) = \mathrm{Concat}([\mathrm{Q}(\mathrm{X}_{\mathrm{V_{g,q_n-1}}}), \ldots, \mathrm{Q}(\mathrm{X}_{\mathrm{K_{g,1}}}), \mathrm{Q}(\mathrm{X}_{\mathrm{V_r}})], \mathrm{dim} = \mathrm{token})$

$X_{V_r} \leftarrow X_{V_r}[-S :]$

**end**

$A = \mathrm{Concat}([\mathrm{t_Q Q}(\mathrm{X}_{\mathrm{K_g}})^{\mathrm{T}}, \mathrm{t_Q X}_{\mathrm{K_r}}^{\mathrm{T}}], \mathrm{dim} = \mathrm{token})$

$A_g = \mathrm{Softmax}(\mathrm{A})[: -\mathrm{S}], A_r = \mathrm{Softmax}(\mathrm{A})[-\mathrm{S} :]$

$t_O = A_g Q(X_{V_g}) + A_r X_{V_r}$

KV cache $\leftarrow Q(X_{K_g}), X_{K_r}, Q(X_{V_g}), X_{V_r}$

**return** $t_O$

**end**

**function** Kquant($X_K \in \Re^{l \times d}$):

$r = l \% S$

$X_{K_g} = X_K[: l - r],\ X_{K_r} = X_K[l - r :]$

$s_n = l // S$

**if** $s_n \leq (q_n - 1)$:

$X_{K_g,i} = X_K[(l - r) - (i \times S + : (l - r) - (i - 1) \times S]$ for $i$ in range[1:$s_n$]

**else**:

$X_{K_g,i} = X_K[(l - r) - i \times S : (l - r) - (i - 1) \times S]$ for $i$ in range[1:$q_n - 2$]

$X_{K_g,q_n-1} = X_K[: (l - r) - (q_n - 2) \times S]$

**if** $s_n \leq (q_n - 1)$:

$Q(X_{K_{g,i}})$ = GrpQuant($X_{K_g,i}$, dim = channel, qbits = $B_{i+1}$, numGroup = $S//G$) for in range[1:$s_n$]

**else**:

$Q(X_{K_{g,i}})$ = GrpQuant($X_{K_g,i}$, dim = channel, qbits = $B_{i+1}$, numGroup = $S//G$) for in range[1:$q_n - 2$]

$Q(X_{K_{g,q_{n-1}}})$ = GrpQuant($X_{K_g,q_{n-1}}$, dim = channel, qbits = $B_{q_n}$, numGroup =

$(l - S \times (q_n - 2) - r)//G)$

**if** $s_n \leq (q_n - 1)$:

$Q(X_{K_g}) = \mathrm{Concat}([\mathrm{Q}(\mathrm{X}_{\mathrm{K}_{\mathrm{g},1}}), \ldots, \mathrm{Q}(\mathrm{X}_{\mathrm{K}_{\mathrm{g},\mathrm{s}_\mathrm{n}}})], \mathrm{dim} = \mathrm{token})$

**else**:

$Q(X_{K_g}) = \mathrm{Concat}([\mathrm{Q}(\mathrm{X}_{\mathrm{K}_{\mathrm{g},1}}), \ldots, \mathrm{Q}(\mathrm{X}_{\mathrm{K}_{\mathrm{g},\mathrm{q}_{\mathrm{n}-1}}})], \mathrm{dim} = \mathrm{token})$

**return** $Q(X_{K_g}), X_{K_r}$

**end**